\documentclass{ieeeaccess}
\usepackage{cite}
\usepackage{amsmath,amssymb,amsfonts}
\usepackage{algorithmic}
\usepackage{graphicx}
\usepackage{textcomp}

\usepackage{bm}
\usepackage{multirow}
\usepackage{array}
\usepackage{hyperref}
\def\BibTeX{{\rm B\kern-.05em{\sc i\kern-.025em b}\kern-.08em
    T\kern-.1667em\lower.7ex\hbox{E}\kern-.125emX}}

\begin{document}
\history{Date of publication xxxx 00, 0000, date of current version xxxx 00, 0000.}
\doi{10.1109/ACCESS.2019.DOI}

\title{Learning a Word-Level Language Model with Sentence-Level Noise Contrastive Estimation for Contextual Sentence Probability Estimation}
\author{\uppercase{Heewoong Park, Sukhyun Cho, and Jonghun Park}}
\address{Department of Industrial Engineering \& Center for Superintelligence,
Seoul National University, Seoul 08826, Republic of Korea}
\tfootnote{This work was supported by Kakao and Kakao Brain corporations, and in part by the National Research Foundation of Korea (NRF) grant funded by the Korea government (MSIT) (No. NRF-2019R1F1A1053366). The authors thank the administrative support from the Institute for Industrial Systems Innovation of Seoul National University.}

\markboth
{H. Park \headeretal: Learning a Word-Level Language Model with Sentence-Level Noise Contrastive Estimation}
{H. Park \headeretal: Learning a Word-Level Language Model with Sentence-Level Noise Contrastive Estimation}

\corresp{Corresponding author: Jonghun Park (e-mail: jonghun@snu.ac.kr).}

\begin{abstract}
Inferring the probability distribution of sentences or word sequences is a key process in natural language processing.
While word-level language models (LMs) have been widely adopted for computing the joint probabilities of word sequences, they have difficulty in capturing a context long enough for sentence probability estimation (SPE).
To overcome this, recent studies introduced training methods using sentence-level noise-contrastive estimation (NCE) with recurrent neural networks (RNNs).
In this work, we attempt to extend it for contextual SPE, which aims to estimate a conditional sentence probability given a previous text.
The proposed NCE samples negative sentences independently of a previous text so that the trained model gives higher probabilities to the sentences that are more consistent with the context.
We apply our method to a simple word-level RNN LM to focus on the effect of the sentence-level NCE training rather than on the network architecture.
The quality of estimation was evaluated against multiple-choice cloze-style questions including both human and automatically generated questions.
The experimental results show that the proposed method improved the SPE quality for the word-level RNN LM.
\end{abstract}


\titlepgskip=-15pt

\maketitle

\section{Introduction}
\label{sec:introduction}
\PARstart{I}{nferring} the probability distribution of sentences or word sequences is one of the core processes in natural language processing (NLP).
Estimating the probability of a sentence with the inferred distribution is useful for many NLP tasks, since a complete sentence, which consists of at least a subject and a main verb to declare a complete thought, is a basic unit of natural language.
For example, by comparing the probabilities of candidate sentences, the most appropriate sentences can be obtained for machine translation, automatic speech recognition (ASR), and various tasks in machine reading comprehension.

Traditionally, the probability of a sentence is derived as the product of conditional probabilities by using a word-level language model (LM), which can be trained much easier than sentence-level probabilistic models.
For a given sentence $s$, a word-level LM can compute the probability \(p(s)=p(w_1, w_2, ..., w_n)=p(w_1)\prod_{i=2}^{n}{p(w_i|w_1, w_2, ..., w_{i-1})}\), where $s$ consists of word sequence $w_1, w_2, ..., w_n$.
A deep learning approach that trains neural LMs by minimizing a word-level cross-entropy (CE) loss between the model predictions and target words has made a remarkable progress in learning word-level LMs mainly due to its decent model capacity and computational efficiency \cite{mikolov2010recurrent}.

The main drawback of typical word-level LMs, however, is that the captured context depends on the limited length of the previous words, often less than five \cite{huang2018whole, daniluk2017Frustratingly}.
This tendency is prone to limiting the ability of the model in comprehending entire sentence contents and structures \cite{le2012measuring}.
To overcome this limitation, sentence-level LMs that directly estimate the probability of an entire sentence without computing the conditional word probabilities have been proposed recently \cite{huang2018whole, wang2018improved}.
These sentence-level LMs have employed noise-contrastive estimation (NCE) that estimates unnormalized density with sampled noise sentences instead of all possible sentences, which are impossible to enumerate.  
By learning from the direct supervision of entire sentences, the sentence-level LMs appeared to capture long-term context, leading to improved performance on rescoring the N-best list in ASR.

Motivated by the success of the sentence-level LMs, we investigate a problem of contextual sentence probability estimation (SPE) with a neural LM trained by applying a sentence-level NCE method.
We define the contextual sentence probability as the conditional sentence probability given a previous text, which is restricted to the immediately preceding sentence in this work, whereas the previous studies \cite{huang2018whole, wang2018improved} did not consider this kind of context.
In real-world application of machine translation and ASR where a series of consecutive sentences is often fed into the systems, accomodation of a sentence's contextual embedding can improve the output quality.
In addition, for natural language understanding (NLU) tasks such as question answering, cloze-style questions, and textual entailment, the contextual inference is useful for comprehending passages consisting of multiple sentences.

To apply NCE for the contextualized settings, we make a modification under which the probabilities of noise sentences and real sentences are approximated by a single neural network instead of two different models, leading to a convenient and computationally efficient implementation.
Specifically, the noise sentences are sampled without considering its context, and hence its probability is approximated by a neural LM without receiving any preceding text data.
Though the use of approximation does not guarantee theoretically-correct estimation, previous work \cite{huang2018whole} found it work well in practice.
With this NCE formulation, the network learns to assign higher probabilities to the sentences conditioned on more appropriate preceding sentences.
In this work, word-level CE loss is extended to the NCE based loss in order to mitigate data sparsity through exploiting both forms of supervision.

We use derivatives of a word-level recurrent neural network (RNN) LM  \cite{mikolov2010recurrent} for implementation of the estimation model of data distribution.
With the simple neural LM, this study attempts to focus on proposing a training objective rather than devising a novel neural architecture for language modeling.
In noise sampling, we suggest two strategies, including batch NCE \cite{oualil2017batch} and editing original sentences with a pre-trained bidirectional LM \cite{devlin2019bert} to obtain realistic fake sentences.

The models are evaluated with cloze-style questions, whose objective is to fill in the blank to obtain a complete consistent passage.
By selecting the sentence having the highest probability when a candidate sentence substitutes the blank, we can obtain the model answers and assess the performance of SPE.
Experiments are conducted on not only natural language data but also symbolic music data.
On natural language data, the proposed method is validated with synthetic cloze questions as well as the SWAG set \cite{zellers2018swag} where human crowd workers were involved in question generation.
The experimental results show that our method enhances the performance compared to that based on only word-level CE loss.

The contributions of this paper are following:
\begin{itemize}
  \item We study how to apply sentence-level NCE to train a neural LM for contextual SPE.
  \item We demonstrate that the quality of the estimation with a word-level RNN is improved when applying the proposed sentence-level NCE.
  \item The proposed method is verified on cloze-style questions, which broadens the potential application area of sentence-level NCE beyond ASR \cite{huang2018whole, wang2018improved}.
\end{itemize}

The remainder of this paper is organized as follows.
First, we offer a briefing on the related work in Section \ref{sec:related}.  
The neural network structure is described in Section \ref{sec:wordrnn}.
Section \ref{sec:snce} introduces our main idea of how to apply sentence-level NCE training.
Section \ref{sec:exp} provides the experiments on cloze questions.
Finally, the last section summarizes and concludes the paper.

\section{Related Work} \label{sec:related}
NCE method that estimates probability density with unnormalized modeling by means of a sampling strategy was presented in \cite{gutmann2010noise}.
To be specific, given sample $x$, the method is designed to maximize the log-ratio between the estimated probability by a learned model, $p_m(x)$, and the probability derived from a noise distribution, $p_n(x)$, if $x$ is a true sample. It minimizes the ratio, otherwise.
In language processing domain, NCE has been applied to learn word embeddings from a large vocabulary by avoiding softmax computation \cite{mnih2013learning}.

Some recent publications have investigated applying NCE method to SPE.
As an earlier study, He \textit{et al.} \cite{he2016training} applied sentence-level NCE to train a bidirectional word-level RNN LM, but their model was even inferior to simple baselines. 
Wang and Ou \cite{wang2018learning} introduced the NCE when training a trans-dimensional random field LM, which is an undirected graphical model where sentences are instances of a collection of random fields.
In \cite{wang2018improved}, the same authors improved the estimation accuracy with different negative sentence sampling and network updating.
Similar to but independent of \cite{wang2018learning, wang2018improved}, Huang \textit{et al.} \cite{huang2018whole} proposed a whole sentence language model, suggesting additional ways of noise sampling, which was particularly advantageous to ASR.
The key difference between our work and the above is that our model takes into account a preceding sentence as a context to estimate the conditional probability, whereas such contextual information was not exploited by them.
Furthermore, we evaluate the estimation performance on machine comprehension questions instead of ASR.

Our method can be considered as an attempt to exploit multiple sources of supervision from differently parsed sequences.
From this perspective, the hierarchical RNN \cite{lin2015hierarchical}, which operates multiple language models that are connected hierarchically, also optimizes multiple objectives with respect to multiple levels similarly to ours. However, their sentence-level model ignores the order of word tokens when deriving the sentence probability. 
Another approach similar to ours is to define sequence probability as summed probability of all possible decompositions through dynamic programming \cite{van2017multiscale, buckman2018neural}.
Yet, its main drawback is a computation burden since it requires a large vocabulary of multi-scale tokens and considers multiple possible sequences that are tokenized differently from each sentence.

In the sense that training a language model as a processing pipeline for solving NLU tasks, our study relates to ongoing research on models for pre-training generalizable language representations.
In \cite{peters2018deep}, the authors suggested to use contextualized embeddings obtained from  a pre-trained LM as input to another task, demonstrating it on challenging NLP tasks.
ULMfit \cite{howard2018universal} and BERT \cite{devlin2019bert} employ an approach that replaces the final layer of a pre-trained LM according to the form of a target task and fine-tunes the network, and in particular the latter \cite{devlin2019bert} showed significantly improved performance on multiple benchmark datasets.
Since our method focuses on SPE, the above fine-tuning methods can be applied equally to ours, which may lead to further improvement.

\section{Word-Level RNN LM for Sentence Probability Estimation} \label{sec:wordrnn}
A word-level RNN LM \cite{mikolov2010recurrent} is a popular deep neural LM that benefits from the strength of RNN in dealing with sequential data.
The network takes as input a word each time step and then outputs the estimation of the probability distribution for the next word.
The estimation of the joint probability over an entire sentence can be naturally obtained by the product of those conditional probabilities.
Due to its popularity and prominence in NLP, the word-level RNN LM has been compared as a baseline model for a variety of NLU tasks \cite{schwartz2017story, park2018word}.
Network architectural modification, including replacing or combining recurrent layers with convolutional layers \cite{kim2016character, hassan2018convolutional, zhang2018combining}, inserting attention mechanism \cite{daniluk2017Frustratingly}, or adopting a transformer-based LM \cite{radford2018improving, yang2019xlnet} can be applied together with our main idea, and it is left for future research.

\Figure[t!]()[width=0.99\columnwidth]{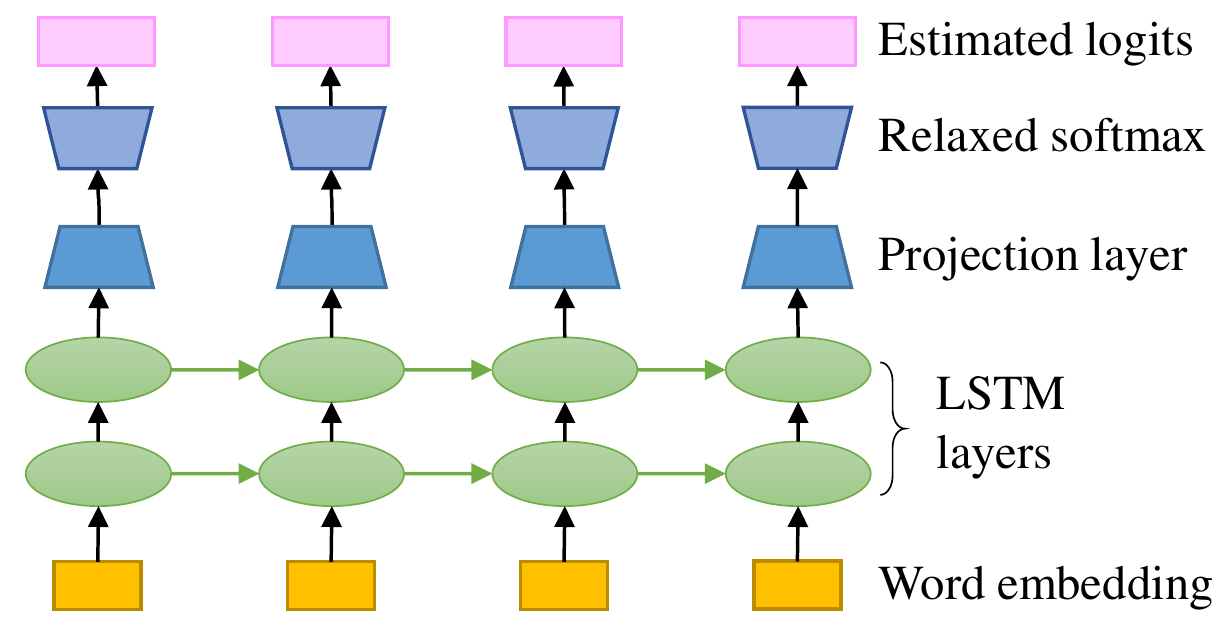}
{Word-level RNN structure.\label{fig:network}}

In detail, we reuse the most part of the structure from \cite{park2018word}, which evaluated word-level RNN LMs for sentence completion, except the facts that input and output embeddings are tied \cite{inan2017tying} and that the last vanilla softmax layer is replaced with a relaxed softmax layer \cite{neumann2018relaxed}, as illustrated in Figure \ref{fig:network}.
Specifically, each word of a given sequence is mapped to a corresponding trainable word embedding vector in the first layer.
Two recurrent layers and a fully-connected layer for linear projection are stacked on the embedding layer.
After the linearly projected vector is multiplied by output embedding matrix, softmax operation yields the probability vector of the next word.

Relaxed softmax \cite{neumann2018relaxed} allows for each output time-step of the network to derive an input-dependent temperature, which adjusts the prediction of the target categorical distribution so that the larger temperature can yield more uniform categorical distribution.
This relaxed softmax layer requires only one more scalar output for predicting the word distribution, which can be realized by adding one more output node to the projection layer.
The operation helps confidence calibration, which aims to adjust the predicted distribution to be associated with how trustworthy the prediction is \cite{guo2017calibration}.
In terms of language modeling, the calibration gives more confidence where the next word is more obvious.
We assume that the relaxed softmax provides more accurate probability estimation for a sentence than it does for a word since the amount of adjustment is accumulated during the joint probability computation.

\section{Training Word-Level RNN LM with Sentence-Level NCE} \label{sec:snce}
\subsection{Training Loss} \label{subsec:loss}
The loss function comprises three terms.
First, we introduce a loss term for our sentence-level NCE.
To be specific, for given preceding sentence $\bm{a}$, let $\bm{b^{real}}$ be the true next sentence and $\bm{b^{neg}_l}$ be the sampled fake sentences $(l=1,...,\nu)$, where $\nu$ is the number of noise samples per true sample.
Under the NCE training framework, the probability that sentence $\bm{b}$ is the next sentence of $\bm{a}$, $p(\bm{b}\textrm{ is the next of }\bm{a})$, is derived as
  \begin{equation}
  p(\bm{b}\textrm{ is the next of }\bm{a}) = \frac{p_m(\bm{b}|\bm{a})}{p_m(\bm{b}|\bm{a})+\nu p_n(\bm{b}|\bm{a})},
  \end{equation}
where $p_m(\bm{b}|\bm{a})$ and $p_n(\bm{b}|\bm{a})$ are the sentence probabilities in case that $\bm{b}$ is from a model distribution and a noise distribution, respectively.

The model distribution, $p_m(\bm{b}|\bm{a})$, is parameterized by the word-level RNN LM described in Section \ref{sec:wordrnn}, whose weights are updated during training.
For conditioning, the final state of the RNN with input $\bm{a}$ serves as the initial state with which the RNN takes $\bm{b}$ as a word sequence input.
For the noise distribution, we sample noise sentences without considering its context $\bm{a}$, reducing $p_n(\bm{b}|\bm{a})$ to $p_n(\bm{b})$.
This simplification reduces computational burden and also allows more flexible choice for the sampling algorithm, which will be discussed in the next subsection.
Although the algorithms suggested in the following subsection are likely to sample realistic sentences,  they do not provide direct sampling probabilities, unlike N-gram LMs.
Thus, we also approximate $p_n(\bm{b})$ by the same word-level RNN, while the information on $\bm{a}$ is completely ignored in estimating $p_n(\bm{b})$.
With a slight abuse of notation, we denote the approximation of $p_n(\bm{b})$ with the word-level RNN as $p_m(\bm{b})$.
As noted earlier, the approximation without exact computation was empirically found to be satisfactory in practice \cite{huang2018whole}.

Consequently, for a true sample and $\nu$ noise samples, the sentence-level NCE loss, $L_{s}$ is defined as
  \begin{equation}
  \begin{split}
  L_{s} = &-\log{p(\bm{b^{real}}\textrm{ is the next of }\bm{a})} \\ &-\sum_{l=1}^{\nu}{\log{(1-p(\bm{b_l^{neg}}\textrm{ is the next of }\bm{a})})},
  \end{split}
  \end{equation}
which performs a logistic regression that discriminates whether a sentence is consistent with preceding sentence $\bm{a}$.

In addition, a loss term for sentence-level classification CE, $L_{c}$, is defined as
\begin{equation}
L_{c} = -\log{\frac{p_m(\bm{b^{real}}|\bm{a})}{p_m(\bm{b^{real}|\bm{a}}) + \sum_{l=1}^{\nu}{p_m(\bm{b_l^{neg}}|\bm{a})}}}
\end{equation}
The loss is computable when a real sentence and $\nu$ noise sentences are provided together within a single batch, which agrees with our sampling strategies.
Our empirical results shows that this auxiliary term helps to boost the optimization.

Finally, we append the negative log-likelihood of a real sentence, which equals to the word-level CE loss, defined as
\begin{equation}
L_{w} = -\log{p_m(\bm{b^{real}})}
\end{equation}
Including this term helps to alleviate overfitting as well as accelerate convergence, in line with the finding in \cite{radford2018improving}.
All three loss terms are weighted and summed with non-negative hyper-parameters $\alpha$, $\beta$, and $\gamma$.
\begin{equation}
L = \alpha L_{w} + \beta L_{s} + \gamma L_{c}
\end{equation}

\subsection{Noise Sampling} \label{subsec:sample}
In this work, the neural network is trained by a stochastic gradient descent based algorithm, and thus the network is updated for multiple iterations each of which the optimizer computes the gradients by using a batch of data instances.
For each batch, we sample $B$ sentence pairs by selecting two successive sentences randomly from the corpus.
Then in the NCE training, $\nu$ noise sentences per real pair are sampled to make fake pairs by replacing the second sentence of the real pair.
We suggest two methods which sample noise sentences independently of given preceding context sentence $\bm{a}$.

The first method adopts the batch NCE \cite{oualil2017batch}, in which words in a batch other than a target word is treated as noise samples, originally designed for optimization with the word-level CE loss.
In our sentence-level NCE case, for each real pair, the second sentences of the other pairs in the same batch are treated as the noise sentences, and accordingly $\nu$ equals to $B-1$.
Since the likelihood that the same sentences exist within a single batch is quite small in contrast to the word-level batch NCE case, we do not take it into account.

The other method, referred to as sentence resampling in this paper, uses a pre-trained word-level bidirectional RNN to resample transformed sentences from original sentences.
To sample a new sentence replaced with multiple new words, we adopt the masked LM training style \cite{devlin2019bert} in which about 15\% of input word tokens are replaced with the \texttt{<MASK>} token, and then the model learns to reconstruct the original word sequences.
Then, the resampling procedure for the noise sampling uses the following operations: input masking, predicting the sequence of word probabilities (identically to the masked LM training so far), and sampling words at the masked positions with the predicted probabilities.
A bidirectional RNN is optimized in advance for the NCE training during which the RNN is not updated.
We remark that we set the network structure of the bidirectional RNN same as that of \cite{park2018word}, which predicts the target word based on both preceding words and following words, except for tying forward embeddings, backward embeddings, and output embeddings.

These two suggested methods sample more realistic sentences than generating sentences from scratch in a word-by-word manner by using a word-level LM, which often prefers short sentences \cite{irie2018investigation}.
Our preliminary experiments for this paper also revealed that the suggested methods yielded better performance than generating sentences with a word-level LM in terms of next sentence prediction task, which also confirms the results of \cite{huang2018whole}.

\section{Experiments} \label{sec:exp}
To evaluate the performance of contextual SPE, we used cloze-style questions whose goal is to pick an appropriate text to fill in the blank when an incomplete passage containing a blank is given.
This kind of questions is used to assess not only student achievement but also machine comprehension, owing to its requiring linguistic proficiency, common knowledge, and logical reasoning.

We prepared the SWAG \cite{zellers2018swag} questions, each of which contains an incomplete passage consisting of a preceding sentence followed by a blank sentence, and multiple candidate sentences for the blank.
To provide more evaluation results on diverse datasets, we also automatically generate cloze-style questions that have the same format as that of the original SWAG questions of which crowd workers verified the validity.
We refer to this question format as sentence-cloze hereafter.
For the synthetic question generation, we reused the batch NCE and the sentence resampling (Section \ref{subsec:sample}).
In the batch NCE case, the candidate sentences of validation questions were kept from being presented in the training phase by partitioning a corpus. 
While the synthetic questions are much easier than the SWAG questions since the training and validation tasks are aligned, it provides the basic proof-of-concept of the proposed sentence-level NCE training.

To solve a sentence-cloze question, the trained network computes a score for each candidate sentence. Then, the sentence with the highest score is selected as the answer.
We experimented two probability estimate based criteria:
\begin{itemize}
  \item Criterion 1: It uses directly the estimate of $\log{p(\bm{b}|\bm{a})}$, namely $\log{p_m(\bm{b}|\bm{a})}$, as a likelihood score that accords with the purpose to assess the estimation quality.
  \item Criterion 2: It uses the estimate of $\log{p(\bm{b}|\bm{a})}-\log{p(\bm{b})}$, computed as $\log{p_m(\bm{b}|\bm{a})}-\log{p_m(\bm{b})}$, which performed well empirically in the story-cloze task \cite{schwartz2017story}.
\end{itemize}
We compare and analyze the sentence-cloze accuracy results for these two criteria.

\subsection{Sentence Probability Estimation} \label{subsec:spe}

First, we built our model using the ground-truth passages of the SWAG dataset and evaluated against synthetic questions as well as the original questions.
Specifically, the 73K SWAG training questions were assigned for model training, while the 20K SWAG validation questions were divided into halves for validation and holdout sets.
Spacy tokenizer\footnote{\url{https://spacy.io/}} tokenized the text to words, and all words were lowercased.
Word tokens that occur more than three times in the training corpus compose vocabulary set, resulting in about 10K in size, while the other infrequent words were replaced with the \texttt{<UNK>} token \cite{inan2017tying}. 

For comparison, we trained the word-level RNNs with either relaxed or vanilla softmax and different $\alpha$, $\beta$, and $\gamma$ while having the fixed embedding size of 200 and the number of hidden nodes per LSTM layer of 600.
We varied $\alpha$, $\beta$, and $\gamma$ among $\{0.1, 1, 10\}$, and show the results of the best combinations.
To remedy unstable initial training, the network weights were initialized by pre-training with $L_w$ only on the same training corpus, and then the proposed sentence-level NCE method was applied.
In the pre-training, the network was updated during 30 epochs with batch size 20.
We updated this network 20 epochs further to make it fully converged. Then, we used it as a baseline based on the word-level CE loss only.    

In the sentence-level NCE training phase, batch size and the number of training epochs were set to 16 and 50, respectively.
The noise sample size, $\nu$, was set to 15 in both batch NCE and sentence resampling.
The bidirectional network that resamples the sentences from masked sentences had also the embedding size of 200 and the number of hidden nodes per LSTM layer of 600 and was optimized during 50 epochs with batch size 20.
During the experimentation of this work, we used Adam optimizer \cite{kingma2014adam} with learning late $10^{-4}$, clipping gradients to refrain the Euclidean norm of the gradients divided by batch size from exceeding 1. 
A dropout layer with rate 0.5 was placed between LSTM layers and between an LSTM layer and a projection layer.

In addition, we compared ESIM \cite{chen2017enhanced}, whose variants attained the best performance on the SWAG set on the release of the dataset \cite{zellers2018swag}.
In fact, ESIM specializes in classifying the relation of a sentence pair into one of given multi-class labels rather than estimating sentence probabilities.
Under the next sentence prediction circumstance, ESIM can be applied as a binary classifier to determine whether a preceding sentence and a candidate subsequent sentence are consistent.
To do so, we modified ESIM to be optimized with binary CE between the true label and $1/(1+\nu \exp(-v))$, where $v$ is the scalar output of the ESIM network.

With this formulation, $v$ can be regarded as a direct estimate of $p_m(\bm{b}|\bm{a})/p_n(\bm{b}|\bm{a})$, when comparing the training objective of the ESIM with $L_s$.
Moreover, since noise sentences are sampled independently of $\bm{a}$ in this work, $v$ approximates the exponentiation of $\log{p_m(\bm{b}|\bm{a})}-\log{p_n(\bm{b})}$ as a result, which corresponds to the score of criterion 2.
Hence, for sentence-cloze, we selected the choice with the highest $v$ as the answer of the ESIM.
For the ESIM network, we set the embedding size to 200, set the number of hidden nodes per LSTM layer to 200, and used a single-layered LSTM for both input encoding and the inference composition \cite{chen2017enhanced}.
Note that increasing the numbers of layers or hidden nodes did not deliver performance gains.
The word embeddings of our ESIM were initialized with those of the pre-trained network optimized with $L_w$ only.
The hyper-parameters for the weight update and the way how the update was carried out were identical to those in the sentence-level NCE training for fair comparisons.

\begin{table*}[t]
\caption{The accuracy results for synthetic sentence-cloze questions based on the passages of the SWAG dataset. The highest value for each column is highlighted in bold.}
\label{tab:synswag}
\centering
\small
  \begin{tabular}{ |l|r|r|r||r|r||r|r||c| } 
  \hline
  \multirow{2}{*}{Network structure} & \multicolumn{3}{c||}{Loss weight} & \multicolumn{2}{c||}{\begin{tabular}[x]{@{}c@{}}Batch NCE\\accuracy\end{tabular}} & \multicolumn{2}{c||}{\begin{tabular}[x]{@{}c@{}}Sentence resampling\\accuracy\end{tabular}} & \multirow{2}{*}{Note} \\ \cline{2-8}
  & $\alpha$ & $\beta$ & $\gamma$ & criterion1 & criterion2 & criterion1 & criterion 2 & \\ 
  \hline    
  Random & N/A & N/A & N/A & 12.5\% & 12.5\% & 12.5\% & 12.5\% & \\
  \hline
  \multirow{4}{*}{\parbox{2.7cm}{Word-level RNN \\ with vanilla softmax}} & 1 & 0 & 0 & 21.6\% & 43.6\% & 38.8\% & 22.9\% & Word-level CE only \\ \cline{2-9}
  & 0.1 & 10 & 0 & 29.1\% & 53.1\% & 42.4\% & 46.4\% & \multirow{3}{*}{Proposed} \\ \cline{2-8}
  & 0.1 & 10 & 0.1 & 33.3\% & 53.5\% & 41.9\% & 46.1\% & \\ \cline{2-8}
  & 0.1 & 1 & 1 & 39.7\% & 47.5\% & 45.1\% & 38.9\% & \\
  \hline
  \multirow{4}{*}{\parbox{2.7cm}{Word-level RNN \\ with relaxed softmax}} & 1 & 0 & 0 & 21.9\% & 46.3\% & 39.4\% & 26.0\% & Word-level CE only \\ \cline{2-9}
  & 0.1 & 10 & 0 & 30.0\% & 55.3\% & 42.7\% & 48.1\% & \multirow{3}{*}{Proposed} \\ \cline{2-8}
  & 0.1 & 10 & 0.1 & 34.2\% & 56.2\% & 43.4\% & \textbf{48.6\%} & \\ \cline{2-8}
  & 0.1 & 1 & 1 & \textbf{42.7\%} & 52.3\% & \textbf{45.8\%} & 46.7\% & \\
  \hline
  ESIM & N/A & 1 & N/A & N/A & \textbf{63.0\%} & N/A & 41.2\% & Not for SPE \\
  \hline
  \end{tabular}
\end{table*}

\begin{table*}[t]
\caption{The accuracy results for the original SWAG questions. The highest value for each column is marked in bold.}
\label{tab:swag}
\centering
\small
  \begin{tabular}{ |l|r|r|r||r|r||r|r||c| } 
  \hline
  \multirow{2}{*}{Network structure} & \multicolumn{3}{c||}{Loss weight} & \multicolumn{2}{c||}{\begin{tabular}[x]{@{}c@{}}Batch NCE\\training\end{tabular}} & \multicolumn{2}{c||}{\begin{tabular}[x]{@{}c@{}}Sentence resampling\\training\end{tabular}} & \multirow{2}{*}{Note} \\ \cline{2-8}
  & $\alpha$ & $\beta$ & $\gamma$ & Validation & Holdout & Validation & Holdout & \\
  \hline    
  Random & N/A & N/A & N/A & 25.0\% & 25.0\% & 25.0\% & 25.0\% & \\
  \hline
  \multirow{4}{*}{\parbox{2.7cm}{Word-level RNN \\ with vanilla softmax}} & 1 & 0 & 0 & 31.1\% & 30.7\% & 31.1\% & 30.7\% & Word-level CE only \\ \cline{2-9}
  & 0.1 & 10 & 0 & 31.2\% & 29.9\% & 32.3\% & 34.1\% & \multirow{3}{*}{Proposed} \\ \cline{2-8}
  & 0.1 & 10 & 0.1 & 32.0\% & 31.4\% & 34.0\% & 34.3\% & \\ \cline{2-8}
  & 0.1 & 1 & 1 & 32.4\% & 31.3\% & 34.3\% & 32.2\% & \\
  \hline
  \multirow{4}{*}{\parbox{2.7cm}{Word-level RNN \\ with relaxed softmax}} & 1 & 0 & 0 & 32.9\% & 32.8\% & 32.9\% & 32.8\% & Word-level CE only \\ \cline{2-9}
  & 0.1 & 10 & 0 & 34.9\% & 34.7\% & \textbf{36.0\%} & 35.0\% & \multirow{3}{*}{Proposed} \\ \cline{2-8}
  & 0.1 & 10 & 0.1 & 35.9\% & 36.2\% & 35.2\% & 35.4\% & \\ \cline{2-8}
  & 0.1 & 1 & 1 & 36.1\% & 35.7\% & 35.3\% & \textbf{35.6\%} & \\
  \hline
  ESIM & N/A & 1 & N/A & \textbf{39.1\%} & \textbf{38.9\%} & 28.4\% & 27.6\% & Not for SPE \\
  \hline
  \end{tabular}
\end{table*}

Table \ref{tab:synswag} presents the accuracy results of the aforementioned methods for synthetic sentence-cloze questions generated by using the passages of the SWAG dataset as a source text.
Each question was generated to have eight choices containing a single true answer, and therefore the accuracy of a dummy classifier that chooses the answer at random is 12.5\%.
The batch NCE accuracy and the sentence resampling accuracy represent the accuracies on synthetic questions generated by the respective sampling methods.
Those values were measured against the models trained with the corresponding noise sampling.
The checkpoints of the networks were saved every five epochs.
The saved network that achieved the best performance on the validation set was evaluated on the holdout set, and then its holdout accuracy was reported.

As expected, applying the sentence-level NCE training consistently improved the performance, compared to minimizing $L_w$ only.
This tendency was observed regardless of whether using relaxed or vanilla softmax, while the relaxed softmax was consistently superior to vanilla softmax.
On the other hand, the ESIM performed better than our proposed models for the batch NCE, while the proposed models outperformed the ESIM for the sentence resampling.
This indicates that a model for SPE may serve as a complementary component for NLU application.
Additionally, we have noticed that when minimizing $L_w$ only, the batch NCE accuracy was higher with criterion 1 than with criterion 2, whereas the sentence resampling accuracy was not.
These observations imply that our SPE method is more beneficial for the questions where the baseline word-level RNN LM performs better with criterion 1, which is a direct measure of SPE quality, than with criterion 2. 

Next, we used the above trained models to solve the original SWAG questions, showing those results in Table \ref{tab:swag}.
The values in the batch NCE training and the sentence resampling training columns represent the accuracies of the models trained with the respective sampling methods.
Similar to the synthetic sentence-cloze evaluation, the best performing checkpointed network on the validation set was selected and used for evaluating the performance on the holdout set.
We had applied both criteria and chose the criterion 2, which outperformed the other.
Since the questions have four alternative choices, the accuracy of dummy classifier is 25\%.
The models that were optimized with $L_w$ only did not involve noise sampling process in training phase, and accordingly their accuracy results are presented in the batch NCE and sentence resampling training columns redundantly in Table \ref{tab:swag}.  

\begin{table*}[t]
\caption{The accuracy results for synthetic cloze questions based on the ABC notation dataset. The highest value for each column is in bold.}
\label{tab:abc}
\centering
\small
  \begin{tabular}{ |l|r|r|r||r|r||r|r||c| } 
  \hline
  \multirow{2}{*}{Network structure} & \multicolumn{3}{c||}{Loss weight} & \multicolumn{2}{c||}{\begin{tabular}[x]{@{}c@{}}Batch NCE\\accuracy\end{tabular}} & \multicolumn{2}{c||}{\begin{tabular}[x]{@{}c@{}}Sentence resampling\\accuracy\end{tabular}} & \multirow{2}{*}{Note} \\ \cline{2-8}
  & $\alpha$ & $\beta$ & $\gamma$ & criterion1 & criterion2 & criterion1 & criterion 2 & \\
  \hline    
  Random & N/A & N/A & N/A & 12.5\% & 12.5\% & 12.5\% & 12.5\% & \\
  \hline
  \multirow{4}{*}{\parbox{2.7cm}{Word-level RNN \\ with vanilla softmax}} & 1 & 0 & 0 & 66.9\% & 78.8\% & 78.2\% & 52.5\% & Word-level CE only \\ \cline{2-9}
  & 0.1 & 10 & 0 & 75.0\% & 89.6\% & 85.3\% & 86.4\% & \multirow{3}{*}{Proposed} \\ \cline{2-8}
  & 0.1 & 10 & 0.1 & 80.2\% & 89.9\% & 85.3\% & \textbf{86.4\%} & \\ \cline{2-8}
  & 0.1 & 1 & 1 & \textbf{86.9\%} & 88.2\% & \textbf{88.9\%} & 82.8\% & \\
  \hline
  \multirow{4}{*}{\parbox{2.7cm}{Word-level RNN \\ with relaxed softmax}} & 1 & 0 & 0 & 64.2\% & 75.9\% & 76.3\% & 53.4\% & Word-level CE only \\ \cline{2-9}
  & 0.1 & 10 & 0 & 75.1\% & 89.8\% & 85.1\% & 85.7\% & \multirow{3}{*}{Proposed} \\ \cline{2-8}
  & 0.1 & 10 & 0.1 & 80.2\% & 90.2\% & 85.4\% & 85.8\% & \\ \cline{2-8}
  & 0.1 & 1 & 1 & 86.7\% & 88.7\% & 88.7\% & 83.9\% & \\
  \hline
  ESIM & N/A & 1 & N/A & N/A & \textbf{90.9\%} & N/A & 79.5\% & Not for SPE \\
  \hline
  \end{tabular}
\end{table*}

The results confirm that the proposed method improved the accuracy of the word-level RNN on the original SWAG questions, even though it did not exploit the negative sentences humans provided.
In addition, the performance gaps between when using the relaxed softmax and when using the vanilla softmax for the original SWAG questions were larger than those for the synthetic sentence-cloze ones.
On the other hand, the performance of the ESIM was not as good as reported in \cite{zellers2018swag}, which can be explained by the facts that any pre-trained word embeddings learned from external data were not used and that the NCE training did not exploit the human-supervised negative sentences.
Another interesting observation is that the performance of the proposed method did not vary a lot between the batch NCE and the sentence resampling, while on the contrary, ESIM performed poorly when trained with the sentence resampling.
This implies to some extent that SPE with the NCE training is robust to how training questions are generated. 

\subsection{Probabilistic Modeling of Symbolic Music} \label{subsec:abc}

In this subsection, we examine how much our proposed NCE training is generally applicable to sequence probability estimation through an experiment on music dataset.
Symbolic music, which is a logical structure containing sound-related events for representing music, is an attractive benchmark data type for sequence modeling \cite{chung2014empirical, jozefowicz2015empirical, berglund2015bidirectional}.
In particular, we tested on ABC notation\footnote{\url{http://abcnotation.com/}} datasets, consisting of musical scores each of which is written in a sequence of alphabets, digits, and some special characters.
To align a music in ABC notation with a natural language, we treated musical tokens as words and segments that are split every four bars as sentences.

We used the ABC notation dataset of \cite{sturm2016music} that contains 23K transcriptions.
For this experiment, the musically tokenized data the authors provided were used, resulting in the vocabulary size of 160.
We partitioned the dataset to 8:1:1 ratio for training, validation, and holdout, preparing 100k sentence pairs where each sentence comprises four bars for the NCE training.
The generation process of evaluation questions, network size, and optimization parameters complied with the experimentation of the previous subsection.

Table \ref{tab:abc} gives the accuracy results for synthetic cloze questions on the ABC notation dataset.
We were able to consistently confirm that the NCE training method enhanced the performance compared to that based on $L_w$ only.
Likewise, the proposed models were superior to the others in the sentence resampling, while being on a par with the ESIM in the batch NCE.
In contrast to the natural language case, it is difficult to say either the relaxed softmax or vanilla softmax outperformed the other.
Except this, the overall tendency was in accordance with that of the experimental results on the natural language dataset, leaving room for further investigation on more diverse types of sequences.

\section{Conclusion}
We have studied how to apply sentence-level NCE to train a neural LM for contextual SPE, which extends previous work where no contextual information has been exploited.
To adapt NCE to the contextual settings, the proposed approach samples noise sentences without considering its context and learns to assign higher probabilities to more coherent sentences.
Experiments on the sentence-cloze questions have shown that the proposed sentence-level
NCE method improved the performance of SPE for the word-level RNN LM.
For future research, we will focus on more scalable NCE training with compelling pre-trained language representations.

\bibliographystyle{IEEEtran}
\bibliography{references.bib}

\EOD

\end{document}